\documentclass[10pt,twocolumn,letterpaper]{article}

\usepackage[pagenumbers]{cvpr} 

\usepackage{cuted}
\usepackage{graphicx}
\usepackage{amsmath}
\usepackage{amssymb}
\usepackage{booktabs}
\usepackage{multirow}
\usepackage[dvipsnames]{xcolor}
\usepackage{pifont}
\newcommand{\cmark}{\ding{51}}%
\newcommand{\xmark}{\ding{55}}%
\newcommand{\mg}{\color{ForestGreen}} 
\usepackage{caption}
\newcommand{\tcr}[1]{\textcolor{black}{#1}}
\usepackage[dvipsnames]{xcolor}

\definecolor{cvprblue}{rgb}{0.21,0.49,0.74}
\usepackage[pagebackref,breaklinks,colorlinks,citecolor=cvprblue]{hyperref}

\usepackage[capitalize]{cleveref}
\crefname{section}{Sec.}{Secs.}
\Crefname{section}{Section}{Sections}
\Crefname{table}{Table}{Tables}
\crefname{table}{Tab.}{Tabs.}

\def\paperID{*****} 
\def\confName{CVPR}
\def\confYear{2024}

\begin{document}

\title{A Simple Strategy for Body Estimation from Partial-View Images}

\author{Yafei Mao, Xuelu Li, Brandon Smith, Jinjin Li, Raja Bala \\
Amazon Inc. \\ 
{\tt \small \{yafeimao, xueluli, smithugh, jinjinli, rajabl\}@amazon.com}}



\maketitle
\begin{abstract}
Virtual try-on and product personalization have become increasingly important in modern online shopping, highlighting the need for accurate body measurement estimation. Although previous research has advanced in estimating 3D body shapes from RGB images, the task is inherently ambiguous as the observed scale of human subjects in the images depends on two unknown factors: capture distance and body dimensions. This ambiguity is particularly pronounced in partial-view scenarios. To address this challenge, we propose a modular and simple height normalization solution. This solution relocates the subject skeleton to the desired position, thereby normalizing the scale and disentangling the relationship between the two variables. Our experimental results demonstrate that integrating this technique into state-of-the-art human mesh reconstruction models significantly enhances partial body measurement estimation. Additionally, we illustrate the applicability of this approach to multi-view settings, showcasing its versatility.
\end{abstract}
\vspace{-0.8em}
\section{Introduction}
\label{sec:intro}

\begin{figure}[t!]
     \centering
         \centering
         \includegraphics[width=0.4\textwidth]{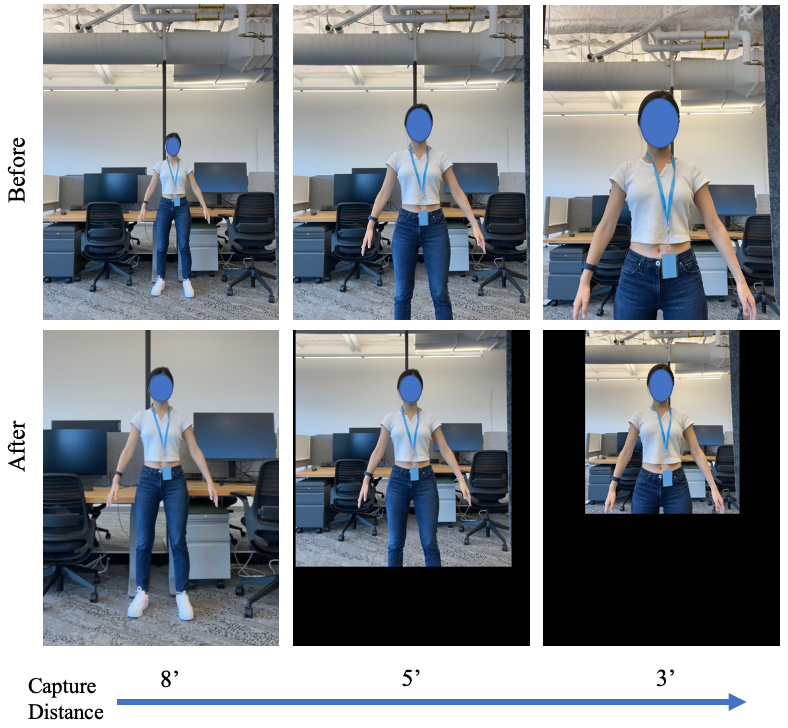}
         \caption{From left to right, the capture distance decreases from 8 to 3 feet. The height normalized versions are shown in the bottom row, where effective capture distance and scale are equalized.}
         \label{fig:height_norm_output}
     \vspace{-0.8em}
\end{figure}

In modern online shopping, obtaining customers' body measurements is crucial for delivering accurate virtual try-on experiences~\cite{pujades2019virtual, mao2022color}. \tcr{A common strategy is to request full-body photos from customers, taken in a designated pose and wearing tight-fitting clothing, along with additional information such as height and/or garment sizes, which are then used to estimate measurements~\cite{walmartweb, zylerweb}}. However, requiring full-body captures can create friction for customers, as it necessitates positioning the camera at a distance of at least 5-6 feet, which may not always be feasible in confined private spaces, as depicted in Fig.~\ref{fig:height_norm_output}. Consequently, customers may occasionally submit partial body (view) images from the knees upwards. 

Previous work~\cite{zhang2021pymaf, joo2021exemplar, li2021hybrik, kolotouros2019learning} has attempted to predict body shapes from monocular images. 
While they can provide reasonable results from full-body inputs, their performance on partial views is unsatisfactory~\cite{rockwell2020full}. 
The difficulty stems from the ill-posed nature of estimating the distance to the camera and the object's size jointly, as the same photo can correspond to multiple valid combinations of the two. When both parameters are unknown, ambiguity arises, such as the inability to determine if the person is close to the camera or has a large body size. 

\begin{figure}
     \centering
         \includegraphics[width=0.45\textwidth]{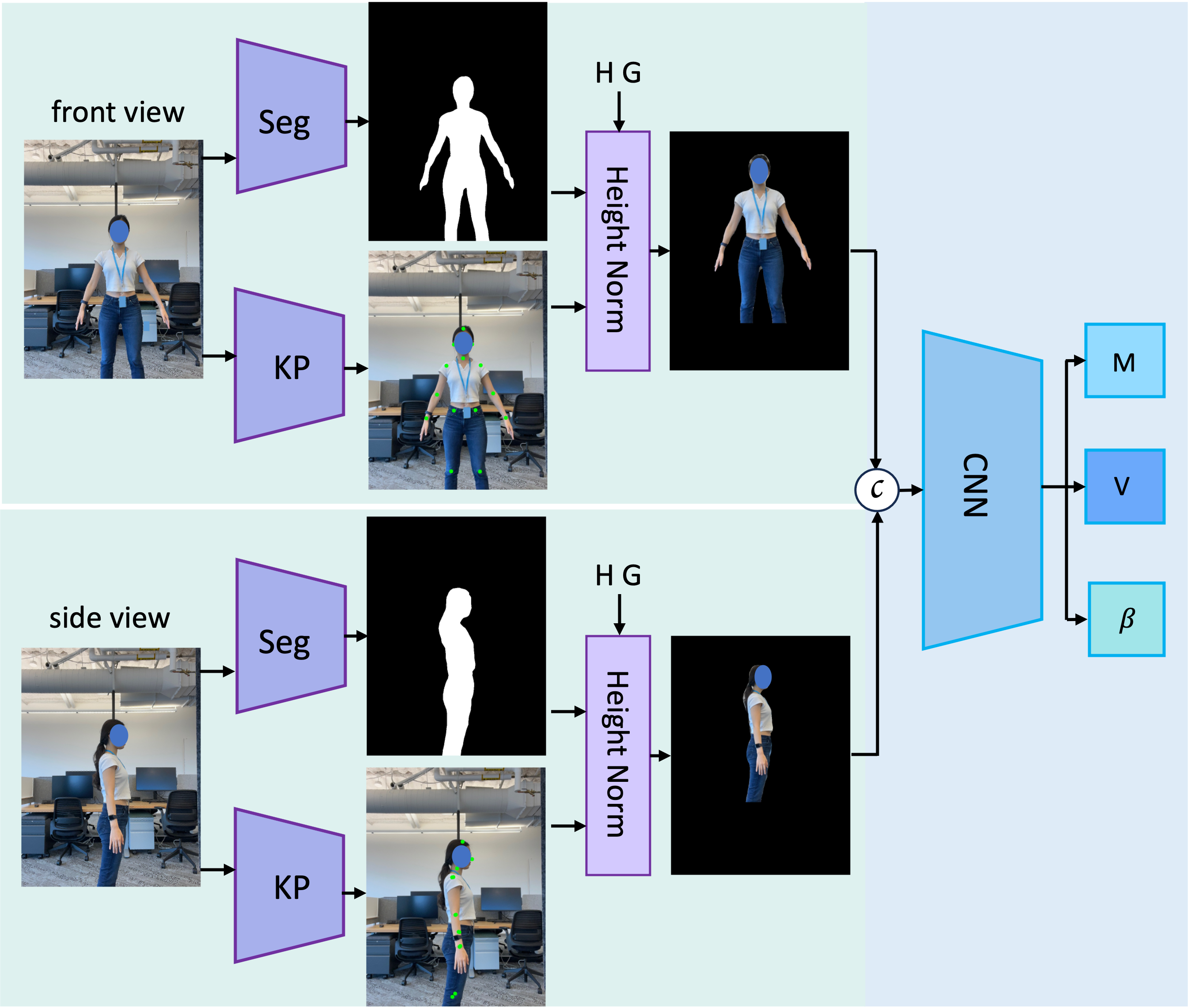}
         \caption{\textbf{Network architecture of BMN model} with height normalization. Segmentation masks and body landmarks are first predicted on each view using the respective models \textit{Seg} and \textit{KP}. Then, this information and the subject's height $H$ and gender $G$ data are passed to the height normalization module to transform the image. Finally, the two views are concatenated together and fed into the network to predict measurements $M$, mesh vertices $V$, and SMPL shape parameters $\beta$.}
         \label{fig:architecture}
  \vspace{-1em}
\end{figure}

In this paper, we address, for the first time, the challenge of body size and capture distance ambiguity in 3D body measurement estimation \tcr{from partial-view images}. We introduce a novel method called \textit{height normalization}, which aims to mitigate these issues by normalizing the scale across all images. Firstly, we estimate a reference 3D skeleton using the subject's gender, height, and the SMPL~\cite{SMPL:2015} model, and then project it to a designated location in the image. Next, utilizing an affine transformation computed from the target skeleton and the detected one, we resize and translate the subject in the images to the designated location. This normalization process allows the network to focus more on learning the characteristics of body dimensions regardless of the capture distance, leading to more accurate measurement predictions. We demonstrate through experiments that our method can be easily integrated into existing HMR models, reducing body estimation errors by up to 2 inches on our real human dataset. Besides, we build a 2-view body estimation pipeline as shown in Fig.~\ref{fig:architecture} to prove its applicability to multi-view settings. 

Our key contributions are: (1) We identify and analyze the challenges related to capture distance and body size ambiguity in estimating body measurements from partial-view images, and propose a straightforward and modular solution. (2) We show that our approach can be easily integrated into existing monocular Human Mesh Recovery (HMR) models and extended to multi-view setups to enhance body measurement accuracy. (3) We conduct thorough experiments to validate the effectiveness of our approach.

\section{Related Work}
\label{sec:relatedwork}
Current methods for obtaining body measurements from RGB images typically involve either regressing from the reconstructed mesh or directly from image features. 

\textbf{Human mesh reconstruction} has received much attention over the years. 
While non-parametric methods directly regress a 3D body representation from images~\cite{Choi_ECCV20,kolotouros2019convolutional, Lin_2021_METRO, Smith_2019_ICCV, Corona2021_SMPLicit,Saito_PIFU_HD_CVPR2020}, parametric methods rely on 3D body models like SMPL\{-X\}~\cite{SMPL:2015,SMPL-X:2019} to estimate the shape and pose parameters via optimization ~\cite{bogo2016keep,SMPL-X:2019,Xiang2019monocular,Zanfir_2021_CVPR} or deep neural regression ~\cite{shapy2022,Biggs2020,hmrKanazawa17,HoloPose2019,rockwell2020full,Kocabas_PARE_2021,Liang_ICCV2019,ExPose:2020,zanfir2021thundr, li2022cliff}. SPIN~\cite{kolotouros2019learning} leverages a collaboration between regression and optimization-based approaches. 
EFT~\cite{joo2021exemplar} augments existing 2D datasets with 3D pseudo-annotations, extreme crop augmentation and auxiliary inputs. PyMAF~\cite{zhang2021pymaf} rectifies predicted parameters through a mesh alignment feedback loop. HyBriK~\cite{li2021hybrik} improves body mesh estimation with inverse kinematics. While these methods achieve reasonable mesh alignment with subjects in full-body images, their performance degrades under partial visibility. Methods like \cite{sengupta2021hierarchical, kolotouros2021probabilistic} predict body shape and/or pose distributions under occlusions. However, the complexity of their dedicated pipeline makes integration into other methods challenging.
 

\textbf{Body measurement estimation} models rely on silhouettes~\cite{dibra2016hs, yan2020silh} and/or synthetic data \cite{ Smith2019TowardsA3} to train deep networks to directly predict measurements. 
Many of these works are hindered by the lack of 3D real data, highlighting the necessity for an efficient data preprocessing strategy. Our proposed method can be easily integrated into the aforementioned pipelines, thus bridging the data gap with minimal effort.

\begin{figure}
     \centering
         \includegraphics[width=0.45\textwidth]{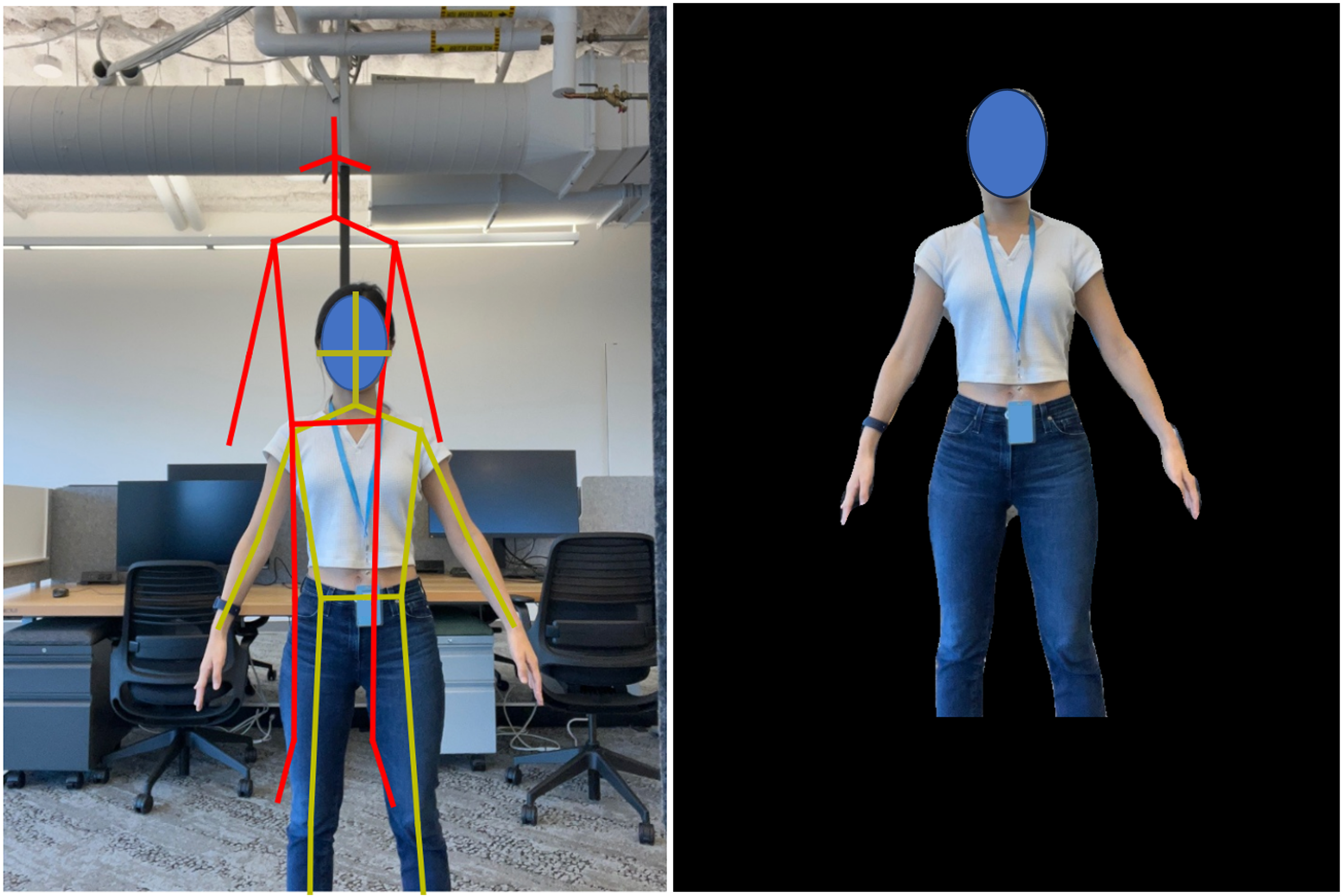}
         \caption{Height normalization involves transforming the skeleton of the person from its detected location (connected by yellow lines) to the designated location (connected by the red lines) through affine transformations.}
         \label{fig:height_norm_method}
  \vspace{-1em}
\end{figure}

\section{Method}\label{sec:method}
Estimating body measurements from partial-view images is challenging. As shown in the first row of Fig.~\ref{fig:height_norm_output}, noticeably different-looking images may correspond to the same set of body shape parameters or measurements, posing a complex one-to-many learning problem for the network. \tcr{Additionally, when dealing with partial-view images, directly estimating depth using camera and human height information is difficult, because the subject's height in the image cannot be easily obtained.} To address these challenges, we propose a height normalization method in this section.

\subsection{Height Normalization}\label{subsec: heightnorm}
Our height normalization strategy aims to position human subjects virtually in the scene at a target skeleton location (as illustrated by the red outline in Fig.~\ref{fig:height_norm_method}). To achieve this, we first generate a reference 3D human mesh using an average SMPL~\cite{SMPL:2015} shape parameters derived from the subject's height and gender in an A-pose. Subsequently, we infer the skeletal joints with the SMPL model and render them onto the scene using a perspective camera. This camera is positioned at a distance of $70$ inches away from the root of the SMPL mesh along the optical axis. This distance was empirically selected to ensure that individuals with a height of less than $78$ inches can be fully projected into the image.

Next, we estimate the 2D body landmarks from the input image using a ResNet50-like keypoint detection model trained on synthetic data (other off-the-shelf keypoint detection models like OpenPose~\cite{cao2019openpose} can also be used, although we will not delve into them in our paper). The detection model outputs the coordinates and the visibility score of the joints, as depicted by the yellow outline in Fig.~\ref{fig:height_norm_method}.  From this information, we derive an affine mapping through least squares fit that aligns the subject's skeletal coordinates with those of the reference body. This transformation scales the partial-view image to a nominal scale and translates the body to center it in the image. We rely on the visible points to compute the transform. An example output of this process is illustrated in Fig.~\ref{fig:height_norm_output}, showing the effective normalization of capture distances.

\subsection{Data Augmentation Strategy}
\tcr{To train models capable of handling both fully visible and partially occluded bodies, we augment} partial views by randomly cropping full-body images from the bottom at various occlusion levels. This process includes deliberately retaining some full-body images without any occlusion (occlusion level $0$). The cropped images are then resized back to their original resolutions. 
The synthesized partial-view data undergoes height normalization before being fed into the network. Additionally, we use segmentation masks to suppress background pixels in the height-normalized images. This step ensures that irrelevant background information doesn't interfere with subsequent body measurement estimation. During test time, all images undergo height normalization directly without random resized cropping.
\begin{table}[t!]
    \centering
     \setlength{\tabcolsep}{2.5pt}
     \resizebox{\columnwidth}{!}{
    \begin{tabular}{l|lllll}
    \toprule
     TP90 (inch)& $\textbf{H}$ & BustGirth & WaistGirth & HipGirth & $\text{Overall}_{59}$ \\
    \midrule
    PyMAF~\cite{zhang2021pymaf} & \xmark & 4.48 & 4.49 & 4.31 & 2.38 \\
    PyMAF~\cite{zhang2021pymaf} & \cmark & 2.92 (\textbf{\mg -1.56}) & 3.67 ({\mg -0.82}) & 4.03 ({\mg -0.28}) & 1.74 ({\mg -0.64}) \\
    HybriK~\cite{li2021hybrik} & \xmark & 4.09 & 4.88 & 5.57 & 2.58 \\
    HybriK~\cite{li2021hybrik} & \cmark & 3.09 ({\mg -1.00})& 3.71 (\textbf{\mg -1.17}) & 3.53 (\textbf{\mg -2.04}) & 1.65 (\textbf{\mg -0.93})\\
    ETF~\cite{joo2021exemplar} & \xmark & 4.02 & 4.58 & 4.27 & 2.40 \\
    ETF~\cite{joo2021exemplar} & \cmark & 3.06 ({\mg -0.96})& 3.68 ({\mg -0.90}) & 3.53 ({\mg -0.74})& 1.66 ({\mg -0.74}) \\
    SPIN~\cite{kolotouros2019learning} & \xmark & 3.82 & 4.51 & 4.17 & 2.41 \\
    SPIN~\cite{kolotouros2019learning} & \cmark & 2.93 ({\mg -0.89}) & 3.79 ({\mg -0.72}) & 3.24 ({\mg -0.93})& 1.67 ({\mg -0.74})\\
    \bottomrule
    \end{tabular}}
    \caption{Comparison of TP90 errors ($\downarrow$) in inches and their differences (in {\mg Green}) on body measurements predicted by SOTA human mesh reconstruction models when trained with and without the height normalization ($\textbf{H}$). The largest difference is in $\mg \textbf{Bold}$.}
    \label{tab:comparison2}
    \vspace{-0.5em}
\end{table}

\section{Results}
\label{sec:Results}
\subsection{Dataset and Metric}
\tcr{Public HMR datasets such as \cite{von2018recovering, ionescu2013human3, mehta2017monocular} do not include information about subjects' height, gender, or body measurements.} Therefore, we collected a training dataset of 6,100 individuals with full-body front-view and left side-view images, along with their height and gender, 59 body measurements, and SMPL-aligned 3D scans. The 59 measurements cover various aspects such as girth, height, and length in areas like the bust, waist, hips, arms, legs, and neck, aiding precise body shape and size estimation. 
Additionally, we obtained at-home photos of 400 new subjects, excluding 3D scans, for our test set. Both training and test images were taken indoors under well-lit conditions, with subjects in an A-Pose wearing form-fitting attire.

Body estimation accuracy is reported in terms of the mean 90th percentile (TP90) errors in inches of the 59 body measurements across the test set subjects to provide a robust statistic of model performance. Additionally, TP90 errors for specific body dimensions are reported for a more granular understanding of the error distribution.

\subsection{Using Height Normalization in HMR Models}

We conducted fine-tuning experiments on state-of-the-art (SOTA) HMR models, including PyMAF~\cite{zhang2021pymaf}, HyBriK~\cite{li2021hybrik}, SPIN~\cite{kolotouros2019learning}, and EFT-HMR~\cite{joo2021exemplar}, both with and without height normalization. 

\tcr{As a standalone module, height normalization can easily serve as a pre-processing step for HMR models to improve body measurement accuracy. During training, all images are processed with height normalization before being passed to the network.} Specifically, utilizing 6,100 frontal view images in our dataset, we first apply resized crop on the input image by uniformly sampling an occlusion level parameter $\alpha$ from between $0$ and $0.4$, \tcr{mimicking scenarios from no occlusion ($0$) to truncation at the waist ($0.4$).} Then we extract segmentation masks using DeepLab-V3~\cite{chen2017rethinking} and detect body landmarks. Subsequently, these partial-view data are height normalized and fed into the network. We incorporate an additional measurement head on top of the feature map from the last layer of the backbone to predict 59 body measurements. This head consists of two linear layers with ReLU activation, and training utilizes the $L_2$ loss function to compare ground truth measurements with predicted ones for supervision.

For all four HMR models, we use a batch size of 64 images, AdamW~\cite{loshchilov2017decoupled} optimizer with a learning rate of $1e-3$, and weight decay of $1e-4$, and trained for $25$ epochs. We present the overall mean TP90 errors of the 59 measurements in inches, along with a detailed analysis of the TP90 errors for Bust Girth, Waist Girth, and Hip Girth. Results are reported based on the $400$ \tcr{images with varying degrees of occlusion}. 
The results are summarized in Table~\ref{tab:comparison2}. It is evident that height normalization leads to a notable reduction in itemized errors, with decreases of up to $2.04$ inches, and an overall error reduction ranging from $0.64$ to $0.93$ inches, thus validating its effectiveness.

\subsection{Using Height Normalization in Multi-View}\label{subsec:blimagetoshape}
\tcr{Next, we demonstrate that the proposed method can also be applied to multiple views.} We developed a simple two-view Body Measurement Network (BMN) baseline model, incorporating the height normalization module separately for each view, as depicted in Fig. \ref{fig:architecture}. We also tested sharing height normalization parameters between the two views and found consistent performance. The height-normalized two-view images are concatenated and forwarded into the CNN backbone and the three task heads for the prediction of body measurements, SMPL mesh vertices, and SMPL shape parameters, respectively. The loss function is a weighted combination of the $L_1$ loss of predicted vertices $V$, shape parameters $\beta$, and measurements $M$: \( \mathcal{L}_\text{total} = w_1 {\mathcal{L}_1}_\text{V} + w_2 {\mathcal{L}_1}_\text{$\beta$} + w_3 {\mathcal{L}_1}_\text{M}\), where $[w_1, w_2, w_3] = [10,1,1]$.


Our BMN baseline uses the ImageNet pretrained ResNet50~\cite{he2016deep} as the backbone and all the task heads comprise two fully connected layers with ReLU activation. The experiments are implemented using PyTorch and are conducted on an NVidia V100 GPU. The model is trained for 75 epochs with a batch size of 8, using the Adam~\cite{kingma2014adam} optimizer with an initial learning rate of $1e-4$. We reduce the learning rate to $0.88\times$ and $0.75\times$ at $112.5K$ and $132K$ steps respectively using the multistep scheduler. 

We compare the BMN baseline model trained with and without height normalization using the 400-subject test set. The mean TP90 error of all 59 measurements and the TP90 error details of five key measurements are summarized in Tables \ref{tab:tp_scores} (a) and (b) respectively, with the best scores highlighted. In (a), height normalization outperforms the baseline across all BMI buckets, resulting in an overall improvement of $0.4$ inches, indicating its effectiveness across diverse body shapes and sizes. In (b), height normalization leads to $0.34$ to $1.27$ inches lower errors in various body parts, demonstrating its effectiveness in disambiguating scale and facilitating training.

\begin{table}[h!]
  \centering
  \small
  \setlength{\tabcolsep}{2pt}
 \begin{subtable}{\linewidth}
    \centering
    \begin{tabular}{@{}lcccccc@{}}
      \toprule
      & \small{Under} & \small{Normal} & \small{Overweight} & \small{ObeseI} & \small{ObeseII} & \small{$\text{Overall}_{59}$} \\
      \midrule
      BMN & 1.67 & 1.44 & 1.38 & 1.77 & 1.97 & 1.52 \\
      BMN\cmark & \textbf{1.23} & \textbf{1.03} & \textbf{0.99} & \textbf{1.39} & \textbf{1.73} & \textbf{1.12} \\
      \bottomrule
    \end{tabular}
    \caption{}
    \label{subtab:tp_scores_a}
  \end{subtable}
  \begin{subtable}{\linewidth}
    \centering
    \begin{tabular}{@{}lcccccc@{}}
      \toprule
      & \small{BustGirth} & \small{WaistGirth} & \small{HipGirth} & \small{BackWidth} & \small{Outseam}\\
      \midrule
      BMN & 2.73 & 3.44 & 2.48 & 1.11 & 1.68 \\
      BMN\cmark & \textbf{1.78} & \textbf{2.17} & \textbf{1.68} & \textbf{0.77} & \textbf{1.24} \\
      \bottomrule
    \end{tabular}
    \caption{}
    \label{subtab:tp_scores_b}
  \end{subtable}
  \caption{TP90 errors ($\downarrow$) in inches for the BMN model w and w/o height normalization: (a) across BMI categories, and (b) for five key measurements. (\cmark) denotes with height normalization.}
  \label{tab:tp_scores}
\end{table}

\begin{figure}[h]
    \centering
    \begin{subfigure}[b]{0.15\textwidth}
        \centering
        \includegraphics[width=\textwidth]{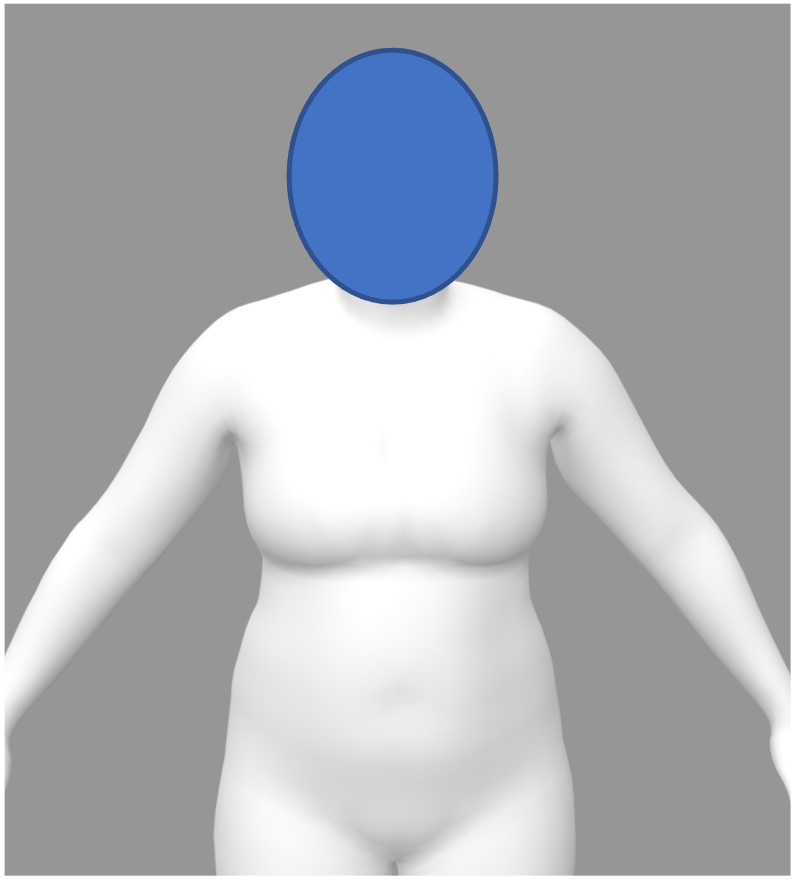}
        \caption{}
        \label{fig:subfigb}
    \end{subfigure}
    \begin{subfigure}[b]{0.15\textwidth}
        \centering
        \includegraphics[width=\textwidth]{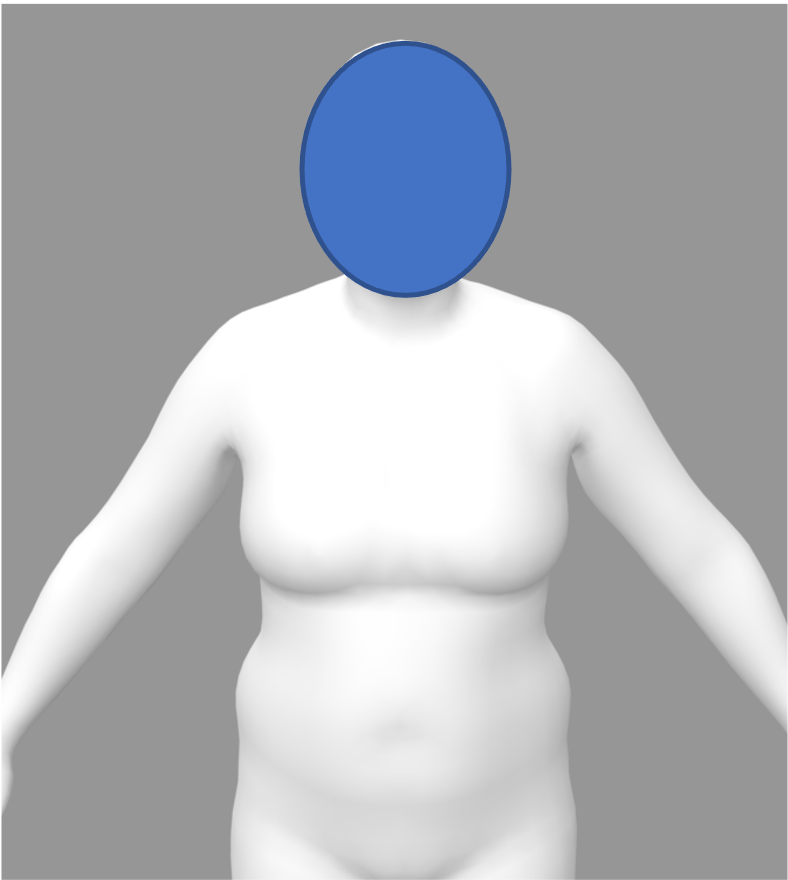}
        \caption{}
        \label{fig:subfigc}
    \end{subfigure}
    \begin{subfigure}[b]{0.15\textwidth}
        \centering
        \includegraphics[width=\textwidth]{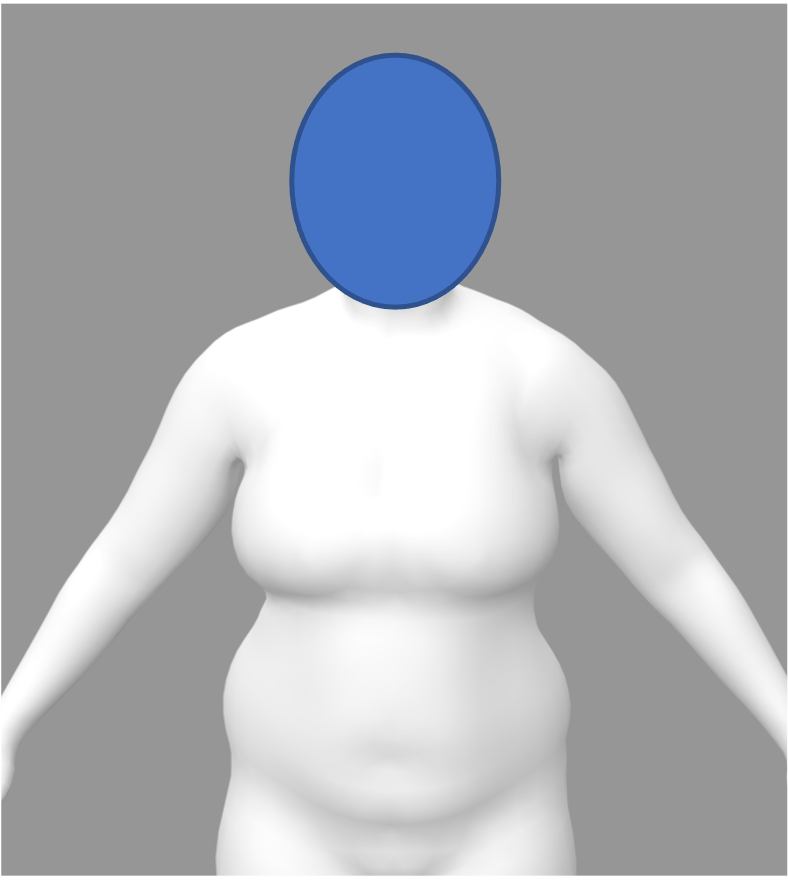}
        \caption{}
        \label{fig:subfigd}
    \end{subfigure}
    \caption{Qualitative comparison of the 3D body mesh predicted by BMN model trained (a) without and (b) with height normalization. (c) shows the SMPL-aligned groundtruth scan of the subject.}
    \label{fig:qualitative}
    \vspace{-0.5em}
\end{figure}

We also qualitatively compare a random test subject’s body mesh predicted by the baseline w and w/o height normalization when the input is a partial-view photo. Our model produces overall better body shape fidelity, especially around the chest and waist areas, shown in Fig. \ref{fig:qualitative}.

\subsection{Ablation Studies}
\textbf{Occlusion level.} During training, we uniformly sample the occlusion level from the range of [0.0, 0.4] for each image. Now, we ablate the lower bound of the hyperparameter occlusion level $\alpha$ on HybriK~\cite{li2021hybrik} in Table~\ref{tab:ablation}~(a). It is evident that our choice of $0.4$ results in the smallest TP90 errors.

\textbf{Background removal.} We ablate the impact of background removal in Table~\ref{tab:ablation}~(b) and find that it reduces TP90 error by $0.09$ inches, providing a marginal benefit.

\begin{table}[h]
  \centering
  \setlength{\tabcolsep}{2pt}
  \begin{minipage}{0.45\linewidth}
    \centering
    \begin{subtable}{\linewidth}
      \centering
      \begin{tabular}{@{}l|cccccc@{}}
        \toprule
        $\alpha$ & 0.5 & 0.4 & 0.3 & 0.2\\
        \midrule
          & 1.76 & \textbf{1.65} & 1.67  & 1.76  \\
        \bottomrule
      \end{tabular}
      \caption{}
      \label{subtab:ablation_a}
    \end{subtable}
  \end{minipage}
  \begin{minipage}{0.45\linewidth}
    \centering
    \begin{subtable}{\linewidth}
      \centering
      \begin{tabular}{@{}l|cc@{}}
        \toprule
        BR & \cmark & \xmark \\
        \midrule
         & \textbf{1.65} &  1.74 \\
        \bottomrule
      \end{tabular}
      \caption{}
      \label{subtab:ablation_b}
    \end{subtable}
  \end{minipage}
  \vspace{-0.7em}
  \caption{Ablation studies on (a) lower bound of occlusion levels $\alpha$ and (b) background removal (BR). We show the overall TP90 errors ($\downarrow$) in inches for both experiments.}
  \label{tab:ablation}
  \vspace{-0.8em}
\end{table}

\section{Conclusions}
We introduce a height normalization technique to address body measurements and scale ambiguity in partial-view body estimation by adjusting the position of human subjects in images to a nominal distance. This method is straightforward yet effective, suitable for both single-view HMR models and multi-view setups. Our experimental results show that integrating this technique enhances body measurement accuracy. However, its applicability may be limited by the requirement for height and gender information. In future work, we plan to explore additional augmentation techniques, such as camera perspective distortion, to further enhance the realism of synthesized partial views.
\label{sec:conclusions}






\newpage
{\small
\bibliographystyle{ieee_fullname}
\bibliography{egbib}
}

\end{document}